\documentclass[runningheads]{llncs}

\usepackage[year=2024,ID=11526]{eccv}

\usepackage{eccvabbrv}

\usepackage{graphicx}
\usepackage{booktabs}

\usepackage[accsupp]{axessibility}  

\usepackage[pagebackref,breaklinks,colorlinks,citecolor=eccvblue]{hyperref}

\usepackage{orcidlink}

\usepackage{amssymb}
\usepackage{amsmath}
\usepackage{makecell}
\usepackage{bbm}
\usepackage{multirow}

\DeclareMathOperator*{\argmin}{arg\,min}

\newcommand{\hanxiR}{\mathbb{R}}
\newcommand{\II}{\text{I}}

\newcommand{\NL}{\\}

\begin{document}

\title{Towards Efficient Pixel Labeling for Industrial Anomaly Detection and Localization} 

\titlerunning{ADClick}


\author{Hanxi Li\inst{1} \and
Jingqi Wu\inst{1}\and
Lin Yuanbo Wu\inst{2}\and
Hao Chen \inst{3} \and \\
Deyin Liu \inst{4} \and
Chunhua Shen \inst{3}
}

\authorrunning{H. Li et al.}

\institute{Jiangxi Normal University, China \and
Swansea University, United Kingdom \and
Zhejiang University, China \and
Anhui University, China
}
\maketitle

\begin{abstract}
  In the realm of practical Anomaly Detection (AD) tasks, manual labeling of anomalous pixels proves to be a costly endeavor. Consequently, many AD methods are crafted as one-class classifiers, tailored for training sets completely devoid of anomalies, ensuring a more cost-effective approach. While some pioneering work has demonstrated heightened AD accuracy by incorporating real anomaly samples in training, this enhancement comes at the price of labor-intensive labeling processes. This paper strikes the balance between AD accuracy and labeling expenses by introducing ADClick, a novel Interactive Image Segmentation (IIS) algorithm. ADClick efficiently generates ``ground-truth'' anomaly masks for real defective images, leveraging innovative residual features and meticulously crafted language prompts. Notably, ADClick showcases a significantly elevated generalization capacity compared to existing state-of-the-art IIS approaches. Functioning as an anomaly labeling tool, ADClick generates
  high-quality anomaly labels (AP $= 94.1\%$ on MVTec AD) based on only $3$ to $5$ manual
  click annotations per training image. Furthermore, we extend the capabilities of ADClick into ADClick-Seg, an enhanced model designed for anomaly detection and localization. By fine-tuning the ADClick-Seg model using the weak labels inferred by ADClick, we establish the state-of-the-art
  performances in supervised AD tasks (AP $= 86.4\%$ on MVTec AD and AP $= 78.4\%$, PRO $=
  98.6\%$ on KSDD2).

\end{abstract}
    
\section{Introduction}
\label{sec:intro}
In contrast to conventional computer vision tasks such as object detection and segmentation, Industrial Anomaly Detection (AD) is tailored for direct application in practical environments. In the context of real-world manufacturing scenarios, anomalous samples are infrequent, and the manual annotation process is prohibitively expensive. Consequently, the majority of AD algorithms focus on developing detection or localization models exclusively based on anomaly-free training samples \cite{defard_padim_2020, bergmann_improving_2019, gudovskiy2022cflow, zhang_pedenet_2021}. While significant progress has been achieved under this constraint \cite{roth2022towards, zavrtanik2021draem}, recent research suggests that the ``one-class'' setting is unnecessarily restrictive. The integration of real defects into the training process has been demonstrated to significantly improve the test accuracy \cite{ding2022catching, li2023efficient, li_target_2023, huang_prototype-based_2023}. 

\begin{figure}[t]  
\centering
\includegraphics [width=0.95\linewidth]{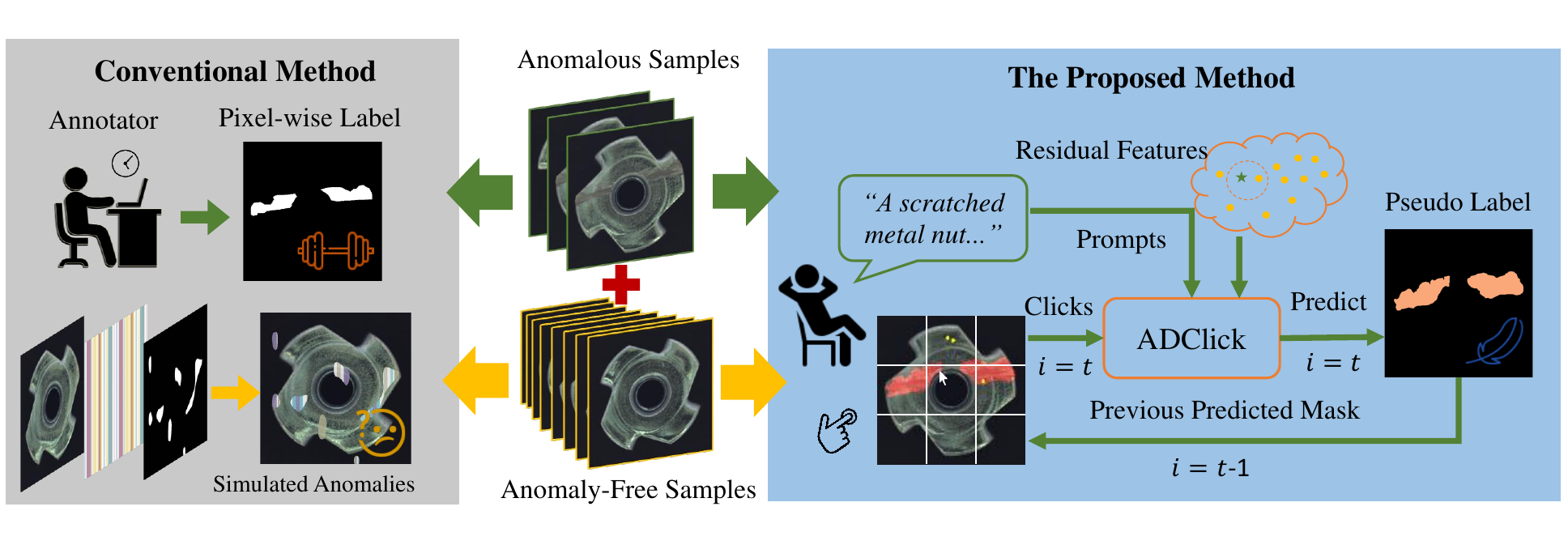}
\caption{
  The illustration of the conventional approach and our proposed approach of label
  generation for anomaly detection and localization. Better viewed in color. 
}\label{fig:main_idea}
\end{figure}

Nowadays, obtaining a few defective samples becomes feasible in modern manufacturing
processes. However, the manual labeling operation is still cost-prohibitive in the construction of Anomaly Detection (AD). Therefore, some AD methods have utilized ``weak labels'' to reduce labeling costs \cite{li2023efficient}. However, these methods overlook the potential of weak labels to
propagate class information to nearby pixels. Inspired by the deployment of ``label propagation'' in various computer vision tasks, such as Image
Matting \cite{xu2017deep, lu2019indices} and Interactive Image Segmentation (IIS)
\cite{wei_focused_2023, zhou_interactive_2023, liu_simpleclick_2023}, this paper proposes to enhance weak labels for anomaly detection or localization using an interactive segmentation approach. Specifically, for a sample image with real defects, the proposed algorithm can generate a high-quality anomaly label mask
with just a few (typically 3 to 5) labeling input clicks into the model. The
core idea of the proposed labeling method, termed ``ADClick'', is illustrated in
Fig.~\ref{fig:main_idea}. Interestingly, ADClick transforms the sparse manual
clicks into dense masks based on three types of input information: human clicks, language
prompts, and the novel residual features. This approach allows for the automatic
generation of more precise mask labels. Extensive experiments on real-world AD benchmarks validate the
superiority of the proposed method.

In summary, 
our main contributions are in three-fold:
 \begin{itemize}
\item
  We introduce a tailored Interactive Image Segmentation (IIS) method dubbed ADClick for  efficient Anomaly Detection (AD) labeling. To our best knowledge, this work is the first to incorporate the IIS concept to enhance weak labels for anomaly detection and localization.
\item
Leveraging the novel ``location-aware'' residual features and defect-specific language prompts,  ADClick surpasses all State-of-the-Art (SOTA) IIS methods by significant margins in label generation. 
   \item
Employing the proposed linguistic features and the IIS-style learning paradigm, the
enhanced ADClick model, termed ADClick-Seg, excels in anomaly detection and   localization. It sets SOTA performances for supervised AD and even beats  the fully-supervised AD methods.
 \end{itemize}

The rest of this paper is structured as follows. In Sec.~\ref{sec:related}, we present
the related algorithm to the proposed one. Sec.~\ref{sec:method} provides a detailed
representation of the proposed ADClick algorithm. The experimental results and analysis
are presented in Sec.~\ref{sec:exp}, and the concluding remarks are outlined in the
last section.


\section{Related Work}
\label{sec:related}

\subsection{Interactive image segmentation}
\label{subsec:iis}

Driven by the huge practical demand of annotation cost reduction, Interactive Image
Segmentation (IIS) has been proposed and improved significantly in recent years. IIS aims to
accurately segmentation the object of interest using various user interaction meanings,
including bounding boxes \cite{zhang2020interactive,xu2017deep}, scribbles
\cite{wu2014milcut}, clicks \cite{xu2016deep} and language prompts
\cite{ding2020phraseclick}. Conventional approaches of interactive image segmentation
\cite{grady2006random,gulshan2010geodesic,kim2010nonparametric,rother2004grabcut}
typically rely on extracting low-level image features and constructing optimization-driven
graphical models. However, these methods often exhibit subpar performance and limited
efficiency. DIOS \cite{xu2016deep} firstly introduces deep learning into the domain of IIS
and inspires the following works
\cite{chen2022focalclick,maninis2018deep,hao2021edgeflow}. The interactive labeling function of this work is designed mainly following the most
recently proposed SimpleClick\cite{liu_simpleclick_2023} algorithm. We generate pseudo labels for anomaly detection which stems essentially from $3$ to $5$ human clicks and thus we term those labels as weak labels. 
\subsection{Referring segmentation}
\label{subsec:referring}

As another recently emerged computer vision task, referring segmentation is typically
cross-modal. The referring segmentation model is supposed to segment the object in the
image according to the linguistic instruction. Various methods are proposed to address
this challenging problem \cite{jain-gandhi-2022-comprehensive, ding2022vlt, yang2022lavt, liu2023polyformer}. 
For instance \cite{jain-gandhi-2022-comprehensive} performs the interactions happening across visual and linguistic modalities and the interactions within each modality simultaneously. Another work in \cite{ding2022vlt} dynamically produces multiple sets of input-specific query vectors and selectively fuses the corresponding responses by these queries. 
Recall that in most anomaly detection tasks, no matter whether in academic challenge or practical
scenarios, the defect category is also given with the training data, one then can introduce the
concept of referring segmentation for cross-modal AD. 

\subsection{Anomaly detection with language guidance}

Recently, some researchers explored to employ Large Language Models
(LLM) for reducing the data dependence within anomaly detection and localization. WinCLIP \cite{jeong2023winclip} pioneers to exploit the features of CLIP model \cite{radford2021learning} for zero- and few-shot anomaly detection. Following WinCLIP \cite{jeong2023winclip}, more subsequent works
\cite{deng_anovl_2023,cao2023segment,gu2023anomalygpt,tamura2023random} further prove the
effectiveness of the language-vision cross-modal features. 

In this paper, we propose to harness the linguistic information mainly following the
LAVT\cite{yang2022lavt} algorithm which integrates linguistic features into visual features via a
pixel-word attention mechanism at each stage of a Swin transformer \cite{liu2021swin} model.


\section{Method}
\label{sec:method}

\subsection{Overview}
\label{subsec:structure}

\begin{figure*}[ht]  
  \centering{
\includegraphics [width=0.95\textwidth]{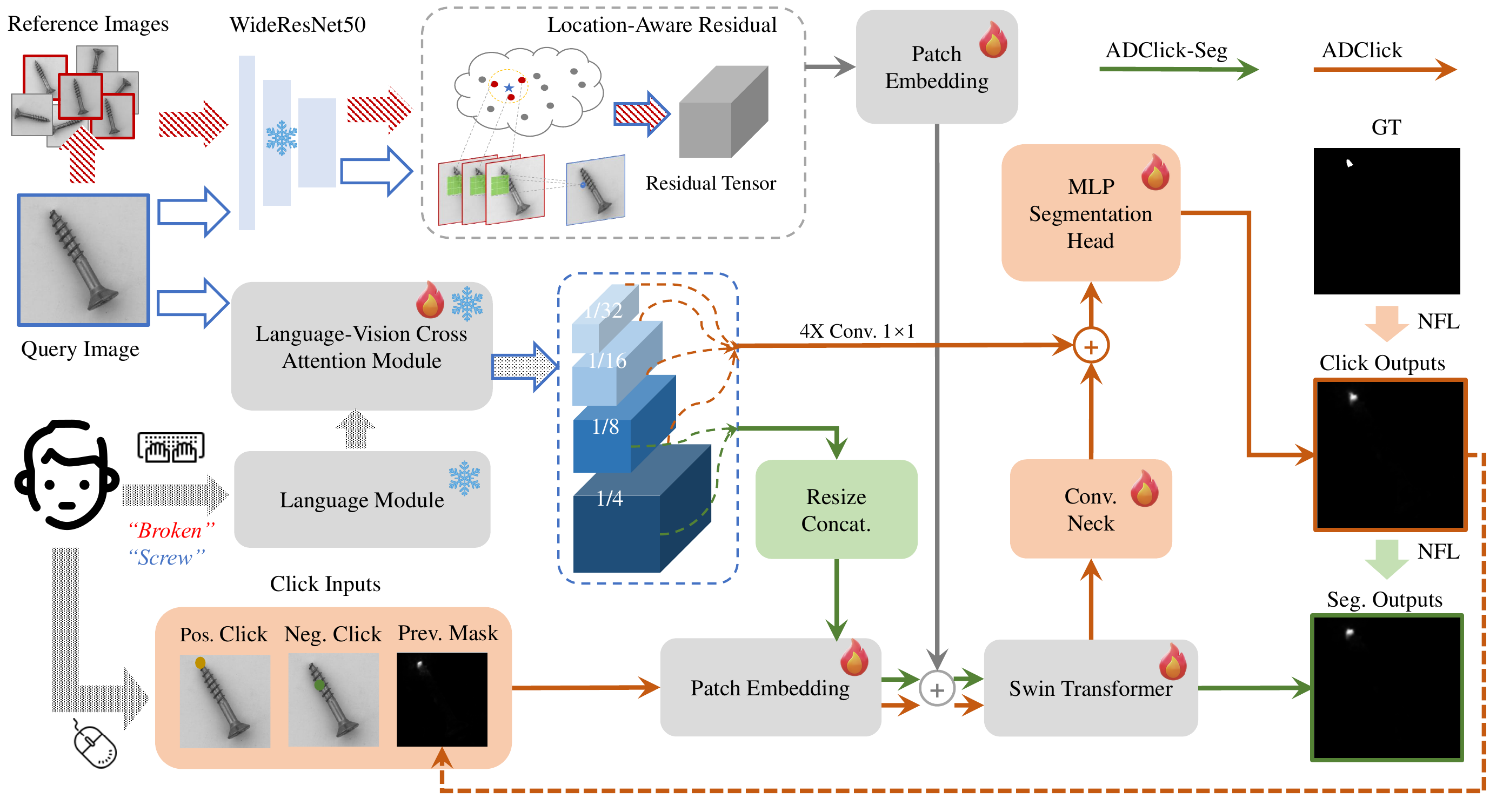}
\caption{
  The illustration of the network structure of the proposed ADClick method. There are four
  main input sources of the model, namely the query image, the reference (defect-free)
  images, the language guidance, and the manual clicks, respectively. Those inputs are
  processed collaboratively as described in this section. Note that the workflows of
  ADClick (in orange) and ADClick-Seg (in green) are slightly different due to the
  different vision tasks and supervision conditions. Better view in color.
}
  \label{fig:method}
}
\end{figure*}

Fig.~\ref{fig:method} depicts the network structure with
different modules of the proposed ADClick method. From the figure, we can see that the
input information of the model backbone is cross-modal. In particular, the
residual tensor is obtained under a global consistency constraint. It is then fed into the
Swin transformer-based backbone to predict the dense pixel labels guided by the
interactive human clicks and the defect-specific language instructions. The click model
and the segmentation-oriented method, namely ADClick-Seg employ a slightly different
module combination and supervision to facilitate the different vision tasks. 

It is worth noting that we do not directly employ the raw-pixel input for this anomaly
labeling task considering that those complex modules in Fig.~\ref{fig:method} could
seriously overfit the small and defect-free training set, as we empirically proved in the
experiment section. In this paper, encouraged by the success of the residual-based anomaly
detection and localization algorithms \cite{roth2022towards, huang_prototype-based_2023,
li2023efficient}, we propose to replace the raw-pixel input $\II_{\text{tst}}$ with the feature
residuals for interactive anomaly labeling. 
The residual-based features are more stable to noises and thus higher generalization
capacity is obtained, as we empirically prove in the experiment. 

Formally, given a normalized input image $\II_{\text{tst}} \in \hanxiR^{H_I\times
W_I\times 3}$ and a successive inputs of human clicks $\mathcal{C} = \{{\bf C}_t
\triangleq [x_t, y_t, \beta_t]^{\text{T}} \mid \forall t = 1, 2, \cdots, T\}$, where
$[x_t, y_t]$ is the x-y coordinate of the clicked pixel and $\beta_t \in \{0, 1\}$
indicating the click is negative (anomaly-free) or positive (anomalous) in the $t$-th
iteration. Then in the annotation iteration $t$, the anomaly mask $\mathrm{M}_t$ is
predicted as follows
\begin{equation}
  \label{equ:click_iteration}
  \forall t, \mathrm{M}_t = \Phi_{\text{Click}}(\mathfrak{R}_{\text{tst}},
  \mathrm{M}_{t-1}, {\bf C}_t,
  \mathrm{V}_{\text{lang}}),
\end{equation}
where the function $\Phi_{\text{Click}}(\cdot)$ denotes the ADClick model for IIS,
$\mathrm{M}_{t-1} \in \hanxiR^{H_I\times W_I}$ stands for the predicted mask in the last
iteration, we store the residual features in the tensor $\mathfrak{R}_{\text{tst}} \in
\hanxiR^{h_f\times w_f \times d_f}$ and the matrix $\mathrm{V}_{\text{lang}} \in
\hanxiR^{Q \times Z}$ is the linguistic feature of the task-guiding sentence containing
$Z$ $Q$-D word features. 

Besides the click model for interactive anomaly annotation, we also designed a variant of
ADClick for anomaly segmentation and term it ``ADClick-Seg''. Even though these two tasks
are similar to each other, the automatic segmentation model is easier to overfit the small
dataset than the human-in-the-loop model. 
Recall that in Fig.~\ref{fig:method}, the MLP-based segmentation head can be viewed as a
complex mask decoder that can easily over-fit the small AD subset, we abandon the
segmentation head in ADClick-Seg. The outputs of the pixel-word module are firstly resized
and concatenated along the channels and then added into the embedded residual features
after a patch-embedding process, as shown in green paths in Fig.~\ref{fig:method}. 

\subsection{Interactive segmentation based on location-aware residual features}
\label{subsec:location_aware}

In general, we adopt the same Swin transformer backbone as the SemiREST algorithm
\cite{li2023efficient}
as it is the best-performing AD method based on residual features. On the other hand, the
most recently proposed CPR algorithm illustrates the merit of using the image-similarity
constraint in patch matching \cite{li_target_2023}. In this paper, the image-similarity is smartly added
into the matching residual generation. Following the terminology of SemiREST
\cite{li2023efficient}, we
firstly extract a set of deep feature vectors from the input image $\II_{\text{tst}}$ as 
\begin{equation}
  \label{equ:extract}
   \Psi_{\mathrm{CNN}}(\II_{\text{tst}}) = \mathfrak{F}_{\text{tst}} \xrightarrow{\texttt{Flatten}}
   \{\mathbf{f}^1_{\text{tst}}, \mathbf{f}^2_{\text{tst}}, \cdots, \mathbf{f}^M_{\text{tst}}\}
\end{equation}
where $\mathfrak{F}_{\text{tst}} \in \hanxiR^{h_f \times w_f \times d_f}$ is the original feature
tensor obtained by the CNN model $\Psi_{\mathrm{CNN}}$ and $\mathbf{f}^j_{\text{tst}} \in
\hanxiR^{d_f}$ denotes the $j$-th element of the $M$ flattened feature vectors. 

Given the anomaly-free reference images $\{\II^1_{\text{ref}}, \II^2_{\text{ref}}, \cdots, \II^R_{\text{ref}}\}$ from
which $N$ ($N = MR$) feature vectors $\mathcal{F}_{\text{ref}} = \{\mathbf{f}^1_{\text{ref}},
\mathbf{f}^2_{\text{ref}}, \cdots, \mathbf{f}^N_{\text{ref}}\}$ can be extracted, let us define the
image index of the feature vectors as $\{\theta^1_{\text{ref}}, \theta^2_{\text{ref}}, \cdots,
\theta^N_{\text{ref}}\}$ with $\theta^i_{\text{ref}} \in [0, R], \forall i = 1, \cdots N$ and the
corresponding x-y coordinates are $\{\boldsymbol{\eta}^1_{\text{ref}}, \boldsymbol{\eta}^2_{\text{ref}},
\cdots, \boldsymbol{\eta}^N_{\text{ref}}\}$ with $\boldsymbol{\eta}^i_{\text{ref}} = [x^i_{\text{ref}},
y^i_{\text{ref}}]^{\text{T}}$.  

For a given test image $\II_{\text{tst}}$, we can calculate its ``global
similarity'' to each reference image $\II^i_{\text{ref}}, \forall i$ using the learning-free metric proposed
in \cite{li_target_2023} 
and store the image indexes of $\II_{\text{tst}}$'s $K$-NN in a set $\Theta_{\text{knn}} =
\{\theta^1_{\text{knn}}, \theta^2_{\text{knn}}, \cdots, \theta^K_{\text{knn}}\}$. Let us define a qualified
reference set for the test feature $\mathbf{f}^j_{\text{tst}}$ as
\begin{equation}
  \label{equ:subset}
  \mathcal{F}_j = \{\forall \mathbf{f}^i_{\text{ref}} \in \mathcal{F}_{\text{ref}} \mid \theta^i_{\text{ref}}
  \in \Theta_{\text{knn}}, \|\boldsymbol{\eta}^j_{\text{tst}} - \boldsymbol{\eta}^i_{\text{ref}}\|_{l_2} <
  \sigma\},
\end{equation}
where $\boldsymbol{\eta}^j_{\text{tst}}$ stands for the the x-y coordinate of
$\mathbf{f}^j_{\text{tst}}$ and $\sigma$ is a predefined small distance parameter. In other words, the test feature can only be matched with the
reference features which are extracted from the $k$-NN images and located at the
neighboring coordinates. In this way, the patch-matching process in this work is
location-aware, especially for those AD datasets with objects \cite{bergmann2019mvtec}. 
Then in mathematics, the proposed location-aware patch-matching writes
\begin{equation}
  \label{equ:constraint_patch_matching}
  \mathbf{f}^{\ast}_j = \argmin_{\mathbf{f}^i_{\text{ref}} \in \mathcal{F}_j}
  \|\mathbf{f}^j_{\text{tst}} - \mathbf{f}^i_{\text{ref}}\|_{l_2}
\end{equation}
Lastly, the residual vector $\mathbf{r}_j$ between $\mathbf{f}^j_{\text{tst}}$ and the retrieved
neighbor $\mathbf{f}^{\ast}_j$ can be calculated in the same way defined in
\cite{li2023efficient}. We
reorganize all the residual vectors into the residual tensor $\mathfrak{R}_{\text{tst}} \in
\hanxiR^{h_f\times w_f \times d_f}$ which is the main input of our ADClick model.

In the literature of AD, the residual features can be categorized into two types, namely
the image-to-image residuals \cite{huang_prototype-based_2023} and the patch-to-patch
residuals \cite{li2023efficient}. The
proposed location-aware residual sits in the middle between these two extremes and thus
better empirical results can be obtained as it is shown in Sec.~\ref{subsec:ad_performance}.

\subsection{Defect specific language prompts}
\label{subsec:language}

Different from the conventional cross-modal AD methods that focus on estimating the
anomaly scores according to the language-vision consistencies. In this paper, on the other
hand, we propose to employ the language prompts in the style of referring image
segmentation \cite{yang2022lavt}. This change stems from the intuition that besides the binary
``perfect vs defective'' representations, some fine-grained linguistic information, such
as the category of the current defect, could also benefit the following anomaly detection
process. Accordingly, we generate the defect-specific language prompts for involving the
useful information. 

\begin{figure}[ht]  
\centering
{
\includegraphics [width=0.95\textwidth]{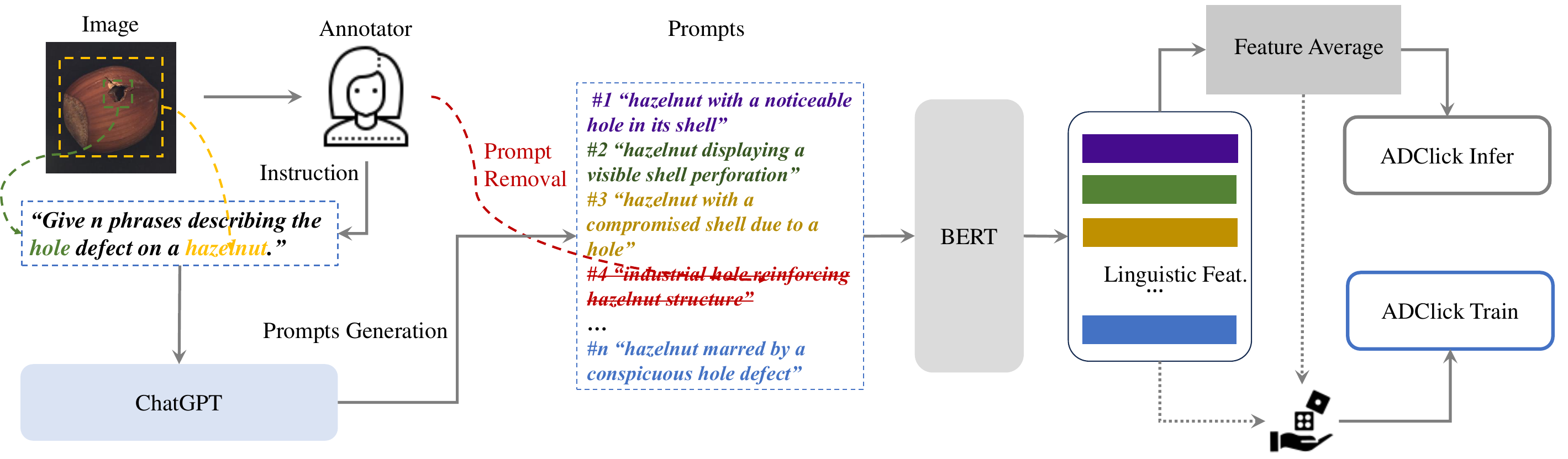}
\caption{
  The generation process of our defect-specific language prompts. The user only needs to supply keywords to form the templatized instruction and then the linguistic features can be generated by using the ChatGPT model, the BERT algorithm, and the final averaging operation. Better view in color.
}
  \label{fig:language}
}
\end{figure}

Fig.~\ref{fig:language} illustrates the generation process of the prompts and the
corresponding linguistic feature obtained by using the popular BERT \cite{devlin2018bert} model.
Instead of generating the linguistic prompts by using the sophisticated templates
\cite{jeong2023winclip, deng_anovl_2023},
we use templatized instructions to control the ChatGPT model \cite{wu2023brief} to automatically
generate multiple ($U$) candidate prompts. In specific, the templated instruction writes
\begin{equation}
  ``\text{Give } n \text{ phrase describing the } \{def\} \text{ defect on a } \{obj\}",
\end{equation}
where $def$ and $obj$ are the category name of the current defect and object,
respectively, $n$ is set as $n > U$ to ensure the sufficient valid prompts. 

As shown in Fig.~\ref{fig:language}, those prompts enjoy more variations than the
predefined description. Note that the annotator also needs to delete the invalid
descriptions occasionally generated by the ChatGPT model, as the strike-through text in
Fig.~\ref{fig:language}. This process could be conducted automatically by training an NLP
classifier but it is out of the scope of this work. 
Mathematically, the prompt feature generation can be described as follows
\begin{equation}
  \label{equ:prompt}
  \mathrm{V}_{\text{lang}} = \frac{1}{U}\sum_{u = 1}^U \Psi_{\text{BERT}}(\phi_u),
\end{equation}
where $\phi_u$ is the $u$-th prompt generated by the ChatGPT model,
$\mathrm{V}_{\text{lang}} \in \hanxiR^{Q\times Z}$ is the average BERT feature with word
embedding dimension $Q$ and maximum $Z$ valid words. Note that this prompt generation can
be swiftly performed before the training and test phase for each specific category of
products and thus it will not lead to any significant drop in algorithm efficiency. 

\begin{figure}[ht]  
\centering
{
\includegraphics [width=0.7\textwidth]{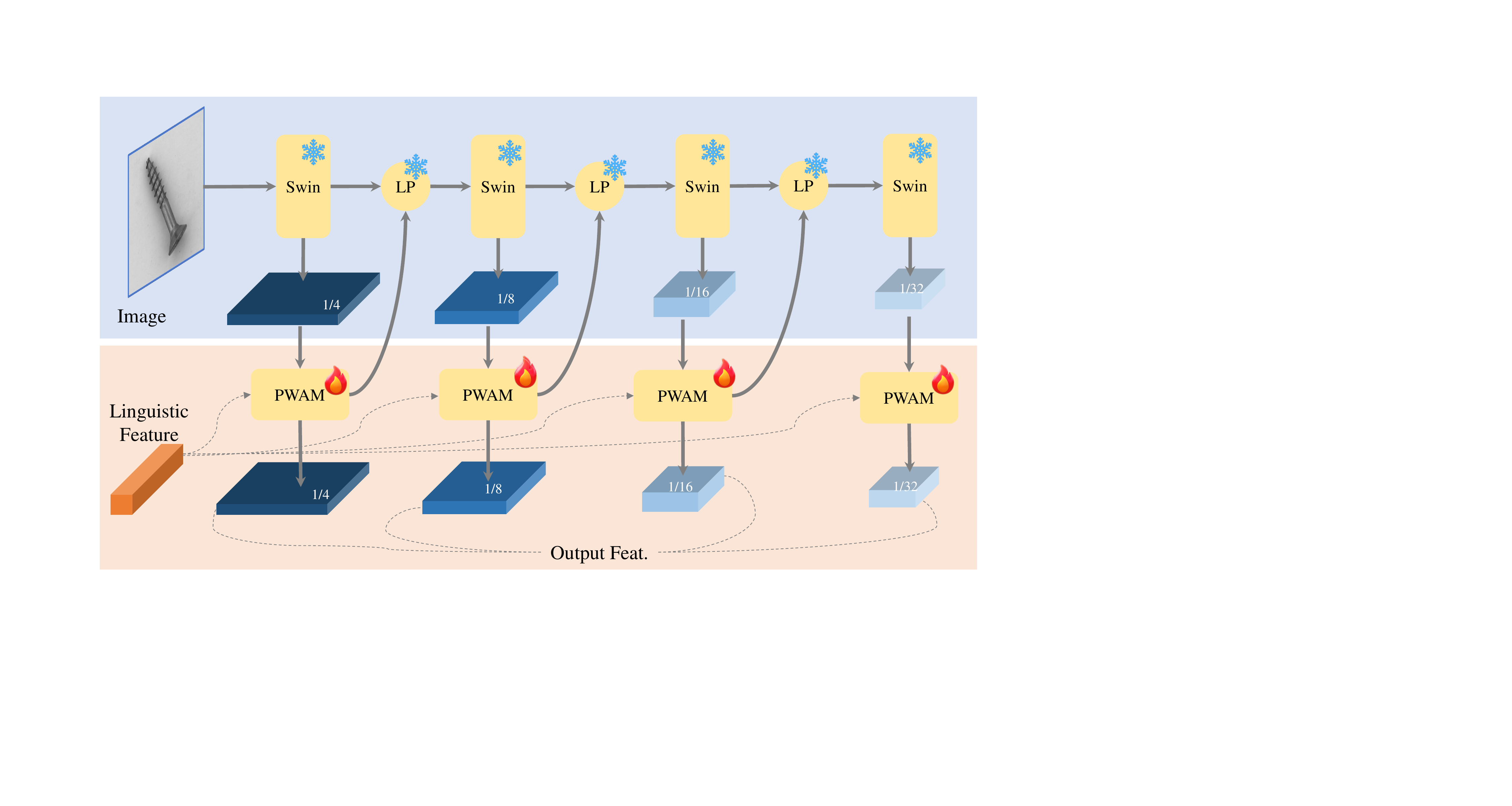}
\caption{
  The Language-Vision Cross Attention Module employed in this paper. Note that the Swin transformer model (in the top row) is fed with images and its parameters are frozen. The PWAM modules (shown in the bottom row) fuse the vision and language features and are fine-tuned during the training stage. 
}
  \label{fig:lavt}
}
\end{figure}

To utilize the information of $\mathrm{V}_{\text{lang}}$, cross attentions are then
performed over $L$ and the corresponding image features within the Pixel-Word Attention
Module (PWAM) proposed in \cite{yang2022lavt}.  Fig.~\ref{fig:lavt} illustrates this
``Language-Vision Cross Attention Module'' employed in this work. Note that the Swin
transformer shown in this figure is different from the backbone of ADClick. It was
pre-trained on a large dataset following the strategy of \cite{yang2022lavt} and we only
fine-tune the PWAMs for AD-related tasks. Finally, the fused cross-modal information is
injected into the ADClick model to complement the residual and click input for better mask
predictions, as shown in Fig.~\ref{fig:method}.

When training, the linguistic features corresponding to the current defect type are fed
into the ADClick and the ADClick-Seg models, with half of the features obtained according
to Eq.~\ref{equ:prompt} and the other half randomly selected from the set
$\{\Psi_{\text{BERT}}(\phi_1), \cdots, \Psi_{\text{BERT}}(\phi_U)\}$. In the inference
stage of ADClick, the averaged linguistic feature $\mathrm{V}_{\text{lang}}$ is input into
the ADClick model for guiding the label generation. On the other hand, as the defect type
is unknown in the test stage of ADClick-Seg, we perform the inference with all the
possible defect types \footnote{For a specific subcategory, the number of defect types is
usually less than $10$ in most popular AD datasets.} and store the corresponding predicted
maps as $\{\mathrm{A}_i \in \hanxiR^{H_I \times W_I} \mid i = 1, 2, \cdots, P\}$. The
pixel value on the coordinate $[x, y]$ of the final anomaly score map $\hat{\mathrm{A}}$
is calculated as
\begin{equation}
    \label{equ:final_score}
    \forall(x, y), ~~\hat{\mathrm{A}}[x, y] = \max^{P}_{i = 1} \mathrm{A}_i[x, y]
\end{equation}

\subsection{The training strategy}
\label{subsec:semi}

We follow the training scheme of most IIS approaches to train our ADClick Model. In
particular, Normalized Focal Loss (NFL) \cite{sofiiuk2019adaptis} is adopted and the model is learned with
the pseudo clicks which are generated by using the same method as SimpleClick \cite{liu_simpleclick_2023}.

As introduced above, the proposed ADClick algorithm is modified to ADClick-Seg for anomaly
segmentation. As a weakly supervised model, our ADClick-Seg model is mainly trained with
the ``pseudo label'' predicted by its prototype, ADClick. Considering the confidence-like
nature of the pixel labels predicted by ADClick, we propose a simple yet effective
semi-supervised strategy to train the segmentation-oriented ADClick model. In specific,
given the ADClick prediction $\mathrm{M} \in \hanxiR^{H_I \times W_I}$, the positive
clicks are stored in the coordinate set $\{\boldsymbol{\alpha}^{+}_u = [x^{+}_u, y^{+}_u]
\mid u = 1, 2, \cdots, C^{+}\}$ while the negative ones form the set
$\{\boldsymbol{\alpha}^{-}_v = [x^{-}_v, y^{-}_v] \mid v = 1, 2, \cdots, C^{-}\}$, we
firstly strengthen the confidence around the click sites (which are more likely to be
correct) as follows
\begin{equation}
    \label{equ:click_strengthen}
    \tilde{\mathrm{M}}(x, y) = 
  \begin{cases}
    1 & \min^{C^{+}}_{u=1}\|\boldsymbol{\alpha}^{+}_u - \boldsymbol{\alpha}\| < d \\
    0 & \min^{C^{-}}_{v=1}\|\boldsymbol{\alpha}^{-}_v - \boldsymbol{\alpha}\| < d \\
    \mathrm{M}(x, y) & else
  \end{cases}    ,
\end{equation}
where $\boldsymbol{\alpha} = [x, y]$ denotes the pixel-coordinate of the prediction map, $d$ is a pre-defined small radius of the ``click discs''. We then generate the tri-valued label map for ADClick-Seg as: 
\begin{equation}
  \label{equ:trimap}
\hat{\mathrm{M}}(x, y) = 
  \begin{cases}
    1 & \tilde{\mathrm{M}}(x, y) > 0.5 + \delta \\
    0 & \tilde{\mathrm{M}}(x, y) < 0.5 - \delta \\
    0.5 & else 
  \end{cases},
\end{equation}
where $\delta$ is a perturbation number randomly selected from $[0, 0.1]$ in each
forward-backward iteration. When training, no loss will be calculated at the $0.5$-valued
pixels on $\hat{\mathrm{M}}$ so that the unconfident part of $\mathrm{M}_T$ will not
influence the model training. The random ``label abandon'' operation introduces more
variation to the training labels and empirically yields more robust segmentation models.
In mathematical terms, the loss of ADClick-Seg writes
\begin{equation}
    L_{\text{NFL}}(x, y) = -  \varepsilon_{x,y} \cdot \frac{|\zeta|^{\gamma}\log
    (1 - |\zeta|)}{\sum_{\tilde{x},\tilde{y}}  \varepsilon_{\tilde{x},\tilde{y}} \cdot|\zeta|^\gamma}, 
\label{eq:NFL}
\end{equation}

where $\varepsilon_{x,y} = \mathbbm{1}(\hat{\mathrm{M}}(x, y) \neq 0.5)$ is the flag variable
turning on/off the loss backpropagation on the site $[x, y]$ and $\zeta =  \hat{\mathrm{M}}(x, y) - \mathrm{A}(x,y)$ denotes the difference between the ADClick-Seg prediction and the pseudo label generated by ADClick. 

Different from most AD methods, we train the ADClick-Seg model mainly in the IIS manner.
Specifically, we employ the Normalized Focal Loss (NFL) \cref{eq:NFL} rather than the
Focal Loss and the ADClick-Seg is also learned with pseudo clicks as the conventional IIS
training process \cite{liu_simpleclick_2023} while the maximum click number is limited to
$3$. 
In the test stage, an all-zero click map is fed into ADClick-Seg to fit the segmentation
scenario. In practice, this step-by-step training strategy brings higher model
generalization capacity.

\subsection{Implementation details}
\label{subsec:detail}
The proposed method employs WideResnet$50$ \cite{zagoruyko2016wide} pre-trained on
ImageNet-1K dataset as feature extractor $\Psi_{\mathrm{CNN}}$. Among these features of
different layers, layer-$1$ features are exploited to calculate the ``global similarity''
according to \cite{li_target_2023}, while layer-$1$, $2$ and $3$ features are resized,
pooled and concatenated together to perform the proposed location-aware patch matching
\cref{equ:constraint_patch_matching} and calculate the residual vectors. Each input image
is resized to $512 \times 512$. As to the location-aware residuals, $K = 50$ global
neighbors are retrieved and $\sigma$ in \cref{equ:subset} is set to $3.2$. 
We apply the foreground mask estimation proposed in \cite{li_target_2023} to the object datasets. 
The BERT
\cite{devlin2018bert} model and the  ``Language-Vision Cross Attention Module'' are
initialized with LAVT \cite{yang2022lavt} weights pre-trained on
RefCOCO\cite{yu2016modeling}. For the linguistic features, $U = 40$ language prompts are
generated by ChatGPT $3.5$ model \cite{wu2023brief} for each type of defect. The AdamW
\cite{loshchilov2018decoupled} optimizer with the learning rate of $3e-5$ and the weight
decay of 0.05 is used and the exponential moving average (EMA)
\cite{haynes2012exponential} trick is adopted.


\section{Experiments}\label{sec:exp}

\subsection{Experimental settings}
\label{subsec:setting}

In this section, we conduct extensive experiments to assess the performance of the proposed ADClick and ADClick-Seg methods, comparing them with state-of-the-art Interactive Image Segmentation (IIS) algorithms and Anomaly Detection (AD) approaches, respectively. For the evaluation of IIS methods, the comparison includes SimplClick \cite{liu_simpleclick_2023}, GPCIS \cite{zhou_interactive_2023}, and the well-known SAM \cite{kirillov_segment_2023}. The ADClick-Seg method, trained with pseudo labels obtained through conventional augmentation or using ADClick with 5 clicks, is compared with 8 SOTA algorithms, namely PatchCore \cite{roth2022towards}, DRAEM \cite{zavrtanik2021draem}, RD \cite{deng2022anomaly}, SSPCAB \cite{ristea2022self}, DeSTSeg \cite{zhang2023destseg}, PyramidFlow \cite{lei2023pyramidflow}, CDO \cite{cao2023collaborative}, SemiREST \cite{li2023efficient}, and the recently proposed CPR \cite{li_target_2023} algorithm.

The experiments are conducted on two well-acknowledged datasets in the Anomaly Detection (AD) field, namely the MVTec AD dataset \cite{bergmann2019mvtec} and the KolektorSDD2 dataset \cite{bovzivc2021mixed}. We evaluate the AD methods using four popular AD metrics: Image-AUROC, Pixel-AUROC, Per-Region Overlap (PRO), and Average Precision (AP). For the involved Interactive Image Segmentation (IIS) methods, conventional mean Intersection over Union (mIoU) scores \cite{zhou_interactive_2023, liu_simpleclick_2023} are also considered. All the aforementioned experiments are executed on a machine equipped with an Intel i5-13600KF CPU, 32GB DDR4 RAM, and an NVIDIA RTX 4090 GPU.

\subsection{Accuracy of label generation}
\label{subsec:label}
To compare the proposed label generation methods, we establish two supervision settings. In the ``pre-trained'' setting, we utilize off-the-shelf SOTA Interactive Image Segmentation (IIS) models, and our ADClick model is pre-trained on all the subcategories of the MVTec AD dataset except the one currently under test\footnote{For the KolektorSDD2 dataset, we pre-train the ADClick model on all the categories of MVTec AD.}. In the ``fine-tuned'' setting, both ADClick and the SOTA IIS algorithms undergo fine-tuning using synthetic anomalous images from the current subcategory generated in the conventional way \cite{li2023efficient}.
As seen in Tab.~\ref{table:iis_mvtec_result}, on MVTec AD, the proposed ADClick model surpasses all the compared SOTA Interactive Image Segmentation (IIS) methods in both 3-click and 5-click tests, across almost all evaluation criteria. The only exception is the mean Intersection over Union (mIoU) numbers in the 5-click test. However, we assert that ADClick does not employ the common ``zoom in'' technique adopted in \cite{liu_simpleclick_2023} and \cite{zhou_interactive_2023}. Improved mIoU performance of ADClick is expected to be achieved with a more sophisticated inference scheme. On the KolektorSDD2 dataset, as indicated in Tab.~\ref{table:iis_mvtec_result}, similar comparison results are observed. These experimental results affirm the superiority of the proposed ADClick method.

\begin{table*}[tb]
\centering
  \caption{The comparison on the AP, PRO, Pixel AUROC and mIoU metrics for interactive segmentation on the MVTec AD and KolektorSDD2 dataset.  $^{\star}$ denotes a model finetuned with simulated anomalies
  and defect-free images of the specific category. }
	\label{table:iis_mvtec_result}
	\resizebox{\linewidth}{!}{
\begin{tabular}{lccccc}
\toprule
  \multirow{3}{*}{Method}&\multicolumn{2}{c}{MVTec AD} && \multicolumn{2}{c}{KolektorSDD2} \\ 
\cmidrule(lr){2-3}\cmidrule(lr){5-6}
               &5-click & 3-click &&5-click & 3-click\\
\cmidrule(lr){2-3}\cmidrule(lr){5-6}
SimpleClick \cite{liu_simpleclick_2023}$_\text{ICCV23}$ & {\color{blue}{\textbf{80.0}}}/{\color{blue}{\textbf{92.2}}}/{\color{blue}{\textbf{97.6}}}/{\color{blue}{\textbf{71.7}}}                                & {\color{blue}{\textbf{58.6}}}/{\color{blue}{\textbf{83.2}}}/{\color{blue}{\textbf{95.1}}}/{\color{blue}{\textbf{53.1}}} 
&& 22.7/78.1/92.9/56.9 & 22.6/79.3/93.2/56.8                                \\

GPCIS \cite{zhou_interactive_2023}$_\text{CVPR23}$     & 75.0/85.3/96.0/58.3                                & 54.8/77.5/93.0/42.7
&&{\color{blue}{\textbf{91.4}}}/{\color{blue}{\textbf{96.4}}}/{\color{blue}{\textbf{98.4}}}/{\color{blue}{\textbf{70.9}}} &{\color{blue}{\textbf{81.6}}}/{\color{blue}{\textbf{92.5}}}/{\color{blue}{\textbf{97.2}}}/{\color{blue}{\textbf{60.5}}} \\
SAM \cite{kirillov_segment_2023}$_\text{ICCV23}$    & 42.7/66.8/89.8/41.5                                & 41.3/64.7/88.7/42.3
& & 25.1/73.7/43.0/30.0 &24.3/75.8/43.9/34.2\\
Ours                   & {\color{red}{\textbf{93.5}}}/{\color{red}{\textbf{98.7}}}/{\color{red}{\textbf{99.7}}}/{\color{red}{\textbf{75.6}}}                                & {\color{red}{\textbf{91.2}}}/{\color{red}{\textbf{98.4}}}/{\color{red}{\textbf{99.6}}}/{\color{red}{\textbf{71.2}}}
&& {\color{red}{\textbf{94.8}}}/{\color{red}{\textbf{99.4}}}/{\color{red}{\textbf{99.8}}}/{\color{red}{\textbf{75.7}}}                                & {\color{red}{\textbf{92.1}}}/{\color{red}{\textbf{99.1}}}/{\color{red}{\textbf{99.6}}}/{\color{red}{\textbf{70.4}}}\\
\cmidrule(lr){2-3}\cmidrule(lr){5-6}
SimpleClick$^{\star}$\cite{liu_simpleclick_2023}$_\text{ICCV23}$ & {\color{blue}{\textbf{92.9}}}/{\color{blue}{\textbf{97.2}}}/{\color{blue}{\textbf{98.6}}}/{\color{red}{\textbf{77.9}}}                                & {\color{blue}{\textbf{87.8}}}/{\color{blue}{\textbf{96.0}}}/{\color{blue}{\textbf{97.7}}}/{\color{blue}{\textbf{70.5}}}
&& {\color{blue}{\textbf{94.6}}}/{\color{blue}{\textbf{98.6}}}/98.8/{\color{red}{\textbf{77.8}}}& {\color{red}{\textbf{91.5}}}/{\color{blue}{\textbf{97.2}}}/{\color{blue}{\textbf{98.5}}}/{\color{red}{\textbf{70.5}}}\\
GPCIS$^{\star}$ \cite{zhou_interactive_2023}$_\text{CVPR23}$      & 86.1/91.5/98.0/65.8                                & 78.0/87.8/95.4/58.0
&&92.8/96.9/{\color{blue}{\textbf{99.0}}}/72.1 & {\color{blue}{\textbf{90.7}}}/93.6/98.4/66.7\\
Ours$^{\star}$                & {\color{red}{\textbf{94.1}}}/{\color{red}{\textbf{98.8}}}/{\color{red}{\textbf{99.7}}}/{\color{blue}{\textbf{77.7}}}&{\color{red}{\textbf{92.1}}}/{\color{red}{\textbf{98.4}}}/{\color{red}{\textbf{99.6}}}/{\color{red}{\textbf{73.7}}}
&& {\color{red}{\textbf{94.9}}}/{\color{red}{\textbf{99.3}}}/{\color{red}{\textbf{99.8}}}/{\color{blue}{\textbf{75.6}}}                             &{\color{red}{\textbf{91.5}}}/{\color{red}{\textbf{98.4}}}/{\color{red}{\textbf{99.5}}}/{\color{blue}{\textbf{68.1}}}\\
\bottomrule                       
\end{tabular}
}

\end{table*}

\begin{table*}[tb]
	\centering
	\caption{The comparison on the Average Precision (AP), Per-Region Overlap (PRO) and Pixel AUROC metrics for unsupervised
anomaly localization on the MVTec AD dataset. The best accuracy in one comparison with the same data and metric condition
is shown in red while
the second one is shown in blue.}
	\label{table:un_result}
	\resizebox{\textwidth}{!}{
		\begin{tabular}{lccccccccccc}
			\toprule
      Category  & \makecell{PatchCore \cite{roth2022towards} \NL (CVPR2022)} &
      \makecell{DRAEM \cite{zavrtanik2021draem} \NL (ICCV2021)}
      & \makecell{RD \cite{deng2022anomaly} \NL (CVPR2022)}                   &
      \makecell{SSPCAB \cite{ristea2022self} \NL (CVPR2022)}
      & \makecell{DeSTSeg \cite{zhang2023destseg} \NL (CVPR2023)}            &
      \makecell{PyramidFlow \cite{lei2023pyramidflow} \NL (CVPR2023)} & \makecell{CDO
      \cite{cao2023collaborative} \NL (TII2023)} & \makecell{SemiREST
      \cite{li2023efficient} \NL (arXiv2023)}                                      &
      \makecell{CPR\cite{li_target_2023}\NL(arXiv2023)} & Ours &  \makecell{Ours \NL (with sliding)}
      \\\hline
			Carpet
			&$64.1$/$95.1$/$99.1$
			&$53.5$/$92.9$/$95.5$         
			&$56.5$/$95.4$/$98.9$                                                  
			&$48.6$/$86.4$/$92.6$                             
			&$72.8$/$93.6$/$96.1$ 
			&$\sim$/$97.2$/$97.4$                                           
			&$53.4$/$96.8$/$99.1$            
			&{\color{red}{$\mathbf{84.2}$}}/{\color{red}{$\mathbf{98.7}$}}/{\color{red}{$\mathbf{99.6}$}} 
		    &$81.2$/$97.6$/$98.9$
		    &$79.3$/$98.4$/$99.3$
		    &{\color{blue}{$\mathbf{82.3}$}}/{\color{blue}{$\mathbf{98.6}$}}/{\color{blue}{$\mathbf{99.4}$}}                         \\
			Grid         
			&$30.9$/$93.6$/$98.8$                                       
			&$65.7$/$98.3$/{\color{red}{$\mathbf{99.7}$}}  
			&$15.8$/$94.2$/$98.3$                                           
			&$57.9$/$98.0$/{\color{blue}{$\mathbf{99.5}$}}                        
			&$61.5$/$96.4$/$99.1$                                           
			&$\sim$/$94.3$/$95.7$                                           
			&$45.3$/$96.1$/$98.4$                                     
			&$65.5$/$97.9$/{\color{blue}{$\mathbf{99.5}$}}                          
			&$64.0$/$97.6$/{\color{blue}{$\mathbf{99.5}$}} 
			&{\color{red}{$\mathbf{73.9}$}}/{\color{blue}{$\mathbf{98.7}$}}/{\color{red}{$\mathbf{99.7}$}}
			&{\color{blue}{$\mathbf{71.9}$}}/{\color{red}{$\mathbf{98.8}$}}/{\color{red}{$\mathbf{99.7}$}}                                          \\
			Leather      
			&$45.9$/$97.2$/$99.3$                                       
			&$75.3$/$97.4$/$98.6$                     
			&$47.6$/$98.2$/$99.4$    
			&$60.7$/$94.0$/$96.3$                                                                          
			&$75.6$/$99.0$/{\color{blue}{$\mathbf{99.7}$}}                        
			&$\sim$/$99.2$/$98.7$                                           
			&$43.6$/$98.3$/$99.2$                                     
			&$79.3$/$99.4$/{\color{red}{$\mathbf{99.8}$}}   
			&$78.5$/{\color{red}{$\mathbf{99.6}$}}/{\color{red}{$\mathbf{99.8}$}}
			&{\color{red}{$\mathbf{80.5}$}}/{\color{red}{$\mathbf{99.6}$}}/{\color{red}{$\mathbf{99.8}$}}
			&{\color{blue}{$\mathbf{79.6}$}}/{\color{blue}{$\mathbf{99.5}$}}/{\color{red}{$\mathbf{99.8}$}} \\
			Tile         
			&$54.9$/$80.2$/$95.7$                                       
			&$92.3$/$98.2$/$99.2$ 
			&$54.1$/$85.6$/$95.7$                                                  
			&{\color{blue}{$\mathbf{96.1}$}}/$98.1$/$99.4$                        
			&$90.0$/$95.5$/$98.0$ 
			&$\sim$/$97.2$/$97.1$                                            
			&$61.8$/$90.5$/$97.2$                                     
			&{\color{red}{$\mathbf{96.4}$}}/{\color{red}{$\mathbf{98.5}$}}/{\color{red}{$\mathbf{99.7}$}}    
			&$94.1$/$98.1$/$99.2$     
			&$95.1$/{\color{blue}{$\mathbf{98.4}$}}/{\color{blue}{$\mathbf{99.5}$}}
			&$93.8$/$98.3$/$99.4$  \\
			Wood         
			&$50.0$/$88.3$/$95.0$                                       
			&$77.7$/$90.3$/$96.4$          
			&$48.3$/$91.4$/$95.8$ 
			&$78.9$/$92.8$/$96.5$                                                                          
			&$81.9$/$96.1$/$97.7$ 
			&$\sim$/{\color{blue}{$\mathbf{97.9}$}}/$97.0$                   
			&$46.3$/$92.9$/$95.8$                                     
			&$79.4$/$96.5$/$97.7$
			&$80.8$/$97.7$/$97.4$
			&{\color{red}{$\mathbf{87.4}$}}/$97.6$/{\color{red}{$\mathbf{98.8}$}}
			&{\color{blue}{$\mathbf{86.7}$}}/{\color{red}{$\mathbf{98.0}$}}/{\color{blue}{$\mathbf{98.7}$}}			
			\\\hline
			Average      
			&$49.2$/$90.9$/$97.6$                                       
			&$72.9$/$95.4$/$97.9$          
			&$44.5$/$93.0$/$97.6$  
			&$68.4$/$93.9$/$96.9$                                                                          
			&$76.4$/$96.1$/$98.1$ 
			&$\sim$/$97.2$/$97.2$                   
			&$50.1$/$96.5$/$98.0$                                     
			&$81.0$/$98.2$/{\color{blue}{$\mathbf{99.3}$}}    
			&$79.7$/$98.2$/$99.0$  
			&{\color{red}{$\mathbf{83.2}$}}/{\color{blue}{$\mathbf{98.5}$}}/{\color{red}{$\mathbf{99.4}$}}
			&{\color{blue}{$\mathbf{82.9}$}}/{\color{red}{$\mathbf{98.6}$}}/{\color{red}{$\mathbf{99.4}$}}
			\\\hline
			Bottle       
			&$77.7$/$94.7$/$98.5$                                       
			&$86.5$/$96.8$/$99.1$          
			&$78.0$/$96.3$/$98.8$                                                  
			&$89.4$/$96.3$/$99.2$           
			&$90.3$/$96.6$/$99.2$                                                 
			&$\sim$/$95.5$/$97.8$                                           
			&$84.1$/$97.2$/$99.3$                                     
			&{\color{red}{$\mathbf{94.1}$}}/{\color{red}{$\mathbf{98.6}$}}/{\color{red}{$\mathbf{99.6}$}}   
			&$92.6$/{\color{blue}{$\mathbf{98.1}$}}/$99.4$
			&$93.3$/$97.7$/$99.4$
			&{\color{blue}{$\mathbf{93.8}$}}/$98.0$/{\color{blue}{$\mathbf{99.5}$}}
			\\
			Cable        
			&$66.3$/$93.2$/$98.4$                                       
			&$52.4$/$81.0$/$94.7$          
			&$52.6$/$94.1$/$97.2$                                                  
			&$52.0$/$80.4$/$95.1$          
			&$60.4$/$86.4$/$97.3$                                                 
			&$\sim$/$90.3$/$91.8$                                           
			&$61.0$/$94.2$/$97.6$                                     
			&$81.1$/$95.3$/$99.1$  
			&$84.4$/$95.2$/{\color{blue}{$\mathbf{99.3}$}}
			&{\color{blue}{$\mathbf{84.5}$}}/{\color{blue}{$\mathbf{95.5}$}}/{\color{red}{$\mathbf{99.4}$}}
			&{\color{red}{$\mathbf{85.4}$}}/{\color{red}{$\mathbf{96.1}$}}/{\color{red}{$\mathbf{99.4}$}}
			\\
			Capsule      
			&$44.7$/$94.8$/$99.0$                                       
			&$49.4$/$82.7$/$94.3$                                                                          &$47.2$/$95.5$/$98.7$                                                  
			&$46.4$/$92.5$/$90.2$                                                                       
			&$56.3$/$94.2$/$99.1$
			&$\sim$/{\color{red}{$\mathbf{98.3}$}}/$98.6$                    
			&$39.5$/$93.0$/$98.6$                                     
			&$57.2$/$96.9$/$98.8$                          
			&$60.4$/$96.3$/{\color{blue}{$\mathbf{99.3}$}}    
			&{\color{blue}{$\mathbf{63.9}$}}/{\color{red}{$\mathbf{97.9}$}}/{\color{red}{$\mathbf{99.4}$}}
			&{\color{red}{$\mathbf{65.6}$}}/$97.8$/{\color{red}{$\mathbf{99.4}$}}
			\\
			Hazelnut     
			&$53.5$/$95.2$/$98.7$                                       
			&{\color{red}{$\mathbf{92.9}$}}/{\color{red}{$\mathbf{98.5}$}}/{\color{red}{$\mathbf{99.7}$}} 
			&$60.7$/$96.9$/$99.0$
			&{\color{red}{$\mathbf{93.4}$}}/{\color{red}{$\mathbf{98.2}$}}/{\color{red}{$\mathbf{99.7}$}}  
			&$88.4$/$97.6$/{\color{blue}{$\mathbf{99.6}$}}                         
			&$\sim$/$98.1$/$98.1$                                           
			&$66.1$/$97.4$/$99.2$                                     
			&$87.8$/$96.1$/{\color{blue}{$\mathbf{99.6}$}} 
			&$88.7$/$97.6$/{\color{blue}{$\mathbf{99.6}$}}    
			&$83.3$/$96.5$/$99.5$
			&$82.1$/$97.2$/$99.5$    \\
			Metal Nut    
			&$86.9$/$94.0$/$98.3$                                       
			&$96.3$/$97.0$/{\color{blue}{$\mathbf{99.5}$}}                         
			&$78.6$/$94.9$/$97.3$                                           
			&$94.7$/$97.7$/$99.4$                         
			&$93.5$/$95.0$/$98.6$                                                 
			&$\sim$/$91.4$/$97.2$                                            
			&$83.8$/$95.7$/$98.5$                                     
			&{\color{blue}{$\mathbf{96.6}$}}/$97.5$/{\color{blue}{$\mathbf{99.5}$}}   
			&$93.5$/$97.5$/$99.3$     
			&{\color{red}{$\mathbf{98.1}$}}/{\color{blue}{$\mathbf{97.8}$}}/{\color{red}{$\mathbf{99.7}$}} 
			&{\color{red}{$\mathbf{98.1}$}}/{\color{red}{$\mathbf{98.0}$}}/{\color{blue}{$\mathbf{99.5}$}}                                               \\
			Pill         
			&$77.9$/$95.0$/$97.8$                                       
			&$48.5$/$88.4$/$97.6$          
			&$76.5$/$96.7$/$98.1$   
			&$48.3$/$89.6$/$97.2$                                                                          
			&$83.1$/$95.3$/$98.7$    
			&$\sim$/$96.1$/$96.1$                                            
			&$81.1$/$96.6$/$98.9$                                     
			&$85.9$/{\color{blue}{$\mathbf{98.4}$}}/{\color{blue}{$\mathbf{99.2}$}}
			&{\color{red}{$\mathbf{91.5}$}}/{\color{red}{$\mathbf{98.7}$}}/{\color{red}{$\mathbf{99.5}$}}    
			&{\color{blue}{$\mathbf{86.2}$}}/$97.2$/$99.0$
			&{\color{blue}{$\mathbf{86.2}$}}/$97.3$/$99.0$
			\\
			Screw        
			&$36.1$/$97.1$/$99.5$              
			&$58.2$/$95.0$/$97.6$          
			&$52.1$/{\color{blue}{$\mathbf{98.5}$}}/{\color{red}{$\mathbf{99.7}$}}   
			&$61.7$/$95.2$/$99.0$          
			&$58.7$/$92.5$/$98.5$                                                 
			&$\sim$/$94.7$/$94.6$
			&$39.4$/$94.3$/$99.0$                                     
			&$65.9$/$97.9$/{\color{red}{$\mathbf{99.7}$}}                            
			&{\color{red}{$\mathbf{71.0}$}}/{\color{red}{$\mathbf{98.7}$}}/{\color{red}{$\mathbf{99.7}$}}  
			&$67.4$/$97.9$/{\color{blue}{$\mathbf{99.6}$}}  
			&{\color{blue}{$\mathbf{68.8}$}}/$98.2$/{\color{blue}{$\mathbf{99.6}$}}  
			\\
			Toothbrush   
			&$38.3$/$89.4$/$98.6$                                       
			&$44.7$/$85.6$/$98.1$          
			&$51.1$/$92.3$/$99.1$    
			&$39.3$/$85.5$/$97.3$                                                                          
			&$75.2$/$94.0$/$99.3$                        
			&$\sim$/{\color{blue}{$\mathbf{97.9}$}}/$98.5$                   
			&$45.9$/$90.5$/$98.9$                                     
			&$74.5$/$96.2$/$99.5$   
			&{\color{red}{$\mathbf{84.1}$}}/{\color{red}{$\mathbf{98.0}$}}/{\color{red}{$\mathbf{99.7}$}}   
			&$77.8$/$97.3$/{\color{blue}{$\mathbf{99.6}$}}  
			&{\color{blue}{$\mathbf{78.2}$}}/$97.3$/{\color{blue}{$\mathbf{99.6}$}}  
			\\
			Transistor   
			&$66.4$/$92.4$/$96.3$                                       
			&$50.7$/$70.4$/$90.9$          
			&$54.1$/$83.3$/$92.3$                                                  
			&$38.1$/$62.5$/$84.8$          
			&$64.8$/$85.7$/$89.1$                                                 
			&$\sim$/$94.7$/{\color{blue}{$\mathbf{96.9}$}}                    
			&$56.3$/$92.6$/$95.3$                                     
			&$79.4$/{\color{blue}{$\mathbf{96.0}$}}/{\color{red}{$\mathbf{98.0}$}}   
			&{\color{red}{$\mathbf{86.7}$}}/{\color{red}{$\mathbf{97.1}$}}/{\color{red}{$\mathbf{98.0}$}}
			&{\color{blue}{$\mathbf{80.9}$}}/$93.1$/$95.5$
 			&{\color{blue}{$\mathbf{80.9}$}}/$94.6$/$96.1$
			\\
			Zipper       
			&$62.8$/$95.8$/$98.9$                                       
			&$81.5$/$96.8$/$98.8$          
			&$57.5$/$95.3$/$98.3$    
			&$76.4$/$95.2$/$98.4$                                                                          
			&$85.2$/$97.4$/$99.1$    
			&$\sim$/$95.4$/$96.6$                                            
			&$55.6$/$94.3$/$98.2$                                     
			&$90.2$/{\color{blue}{$\mathbf{98.9}$}} /{\color{red}{$\mathbf{99.7}$}}    
			&$88.8$/$98.6$/{\color{blue}{$\mathbf{99.6}$}} 
			&{\color{blue}{$\mathbf{90.5}$}}/$98.8$/{\color{red}{$\mathbf{99.7}$}} 
			&{\color{red}{$\mathbf{90.6}$}}/{\color{red}{$\mathbf{99.0}$}}/{\color{red}{$\mathbf{99.7}$}} 
			\\\hline
			Average      
			&$61.1$/$94.2$/$98.4$                                       
			&$66.1$/$89.2$/$97.0$          
			&$60.8$/$94.4$/$97.9$       
			&$64.0$/$89.3$/$96.0$                                                                          
			&$75.6$/$93.5$/$97.5$      
			&$\sim$/$95.2$/$96.6$                                            
			&$61.3$/$94.6$/$98.4$                                     
			&$81.3$/$97.2$/{\color{blue}{$\mathbf{99.3}$}} 
			&{\color{red}{$\mathbf{84.2}$}}/{\color{red}{$\mathbf{97.6}$}}/{\color{red}{$\mathbf{99.4}$}}   
			&$82.6$/$97.0$/$99.1$
			&{\color{blue}{$\mathbf{83.0}$}}/{\color{blue}{$\mathbf{97.4}$}}/$99.2$
			\\\hline
			Total Average 
			&$57.1$/$93.1$/$98.1$                                       
			&$68.4$/$91.3$/$97.3$          
			&$55.4$/$93.9$/$97.8$  
			&$65.5$/$90.8$/$96.3$                                                                          
			&$75.8$/$94.4$/$97.9$ 
			&$\sim$/$95.9$/$96.8$                                            
			&$57.6$/$94.7$/$98.2$                                     
			&$81.2$/{\color{blue}{$\mathbf{97.5}$}}/{\color{red}{$\mathbf{99.3}$}} 
			&$82.7$/{\color{red}{$\mathbf{97.8}$}}/{\color{blue}{$\mathbf{99.2}$}}    
			&{\color{blue}{$\mathbf{82.8}$}}/{\color{blue}{$\mathbf{97.5}$}} /{\color{blue}{$\mathbf{99.2}$}}  
			&{\color{red}{$\mathbf{82.9}$}}/{\color{red}{$\mathbf{97.8}$}}/{\color{blue}{$\mathbf{99.2}$}}  
			\\
			\bottomrule
		\end{tabular}
	}
\end{table*}

\begin{table*}[tb]
	\centering
	\caption{
		Unsupervised anomaly detection (Image AUROC) on MVTec AD \cite{bergmann2019mvtec}. Results are averaged over
		all categories. 	}
	\label{table:un_image_auc_result}
	\resizebox{\linewidth}{!}{
		\begin{tabular}{ccccccccccc}
			\toprule
			\makecell{PatchCore \cite{roth2022towards} \NL (CVPR2022)} & \makecell{DRAEM \cite{zavrtanik2021draem} \NL (ICCV2021)} & \makecell{RD \cite{deng2022anomaly} \NL (CVPR2022)} & \makecell{SSPCAB \cite{ristea2022self} \NL (CVPR2022)} & \makecell{DeSTSeg \cite{zhang2023destseg} \NL (CVPR2023)} & \makecell{CDO \cite{cao2023collaborative} \NL (TII2023)} & \makecell{SimpleNet \cite{Liu_2023_CVPR} \NL (CVPR2023)} &  \makecell{CPR\cite{li_target_2023}\NL(arXiv2023)}                             & \makecell{Ours}  & \makecell{Ours \NL (with sliding)}                      \\ \midrule
			$98.5$                                                     
			&$98.0$                                                    
			&$98.5$                                              
			&$98.9$
			&$98.6$                                                    
			&$96.8$                                                   
			&{\color{blue}{$\mathbf{99.6}$}}                                            
			&{\color{red}{$\mathbf{99.7}$}} 
			&{\color{red}{$\mathbf{99.7}$}} 
			&{\color{red}{$\mathbf{99.7}$}}  \\
			\bottomrule
		\end{tabular}
		 }
\end{table*}

\subsection{Anomaly detection and localization}
\label{subsec:ad_performance}

To assess the effectiveness of the proposed location-aware residual, we compare the ADClick-Seg model with all state-of-the-art (SOTA) methods in the ``unsupervised'' setting. In this setting, only defect-free images are provided for training, and the discriminant models can solely be trained on simulated anomalies \cite{li2023efficient}. Tab.~\ref{table:un_result} illustrates the anomaly localization performances (AP/PRO/Pixel-AUROC), while Tab.~\ref{table:un_image_auc_result} illustrates the anomaly detection performances, specifically the Image-AUROC scores. From the tables, it is evident that the ADClick-Seg model outperforms its prototype SemiRest in most evaluation criteria, with only a marginal ($0.1\%$) gap in Pixel-AUROC. However, when equipped with the sliding-window trick used in SemiREST \cite{li2023efficient}, the proposed ADClick-Seg algorithm achieves a new SOTA AP score of $82.9\%$ and ties with the SOTA CPR method in PRO and Image-AUROC standards. It's important to note that in this ``unsupervised'' setting, no linguistic features are used since the simulated anomalies are challenging to describe within known defect categories.

To evaluate the complete model of ADClick-Seg and verify the ADClick algorithm in practice,  we train the ADClick-Seg model with the pseudo labels generated by ADClick and test the learned AD model comparing with SOTA methods on
the MVTec AD \cite{bergmann2019mvtec} and KolektorSDD2 \cite{bovzivc2021mixed} datasets. For MVTec AD \cite{bergmann2019mvtec}, before training the model on the images of the current subcategory, we pre-train the ADClick-Seg model on all the other subcategories 
Tab.~\ref{table:supervised_mvtec_result} shows the results of anomaly localization (AP/PRO/Pixel-AUROC) on MVTec AD \cite{bergmann2019mvtec} while the anomaly detection results of MVTec AD can be found in Tab.~\ref{table:image_auc_result}.
Tab.~\ref{table:ksdd2_seg_result} reports all the evaluation scores for KolektorSDD2 \cite{bovzivc2021mixed}. From the tables, one can find the obvious performance superiority of the proposed method, even learned with the pseudo labels with a lot of uncertainties. In particular, the new best AP score ($86.4\%$) is reported with the proposed AD algorithm. Furthermore, ADClick-Seg also significantly outperforms all the existing SOTA methods on the KolektorSDD2 \cite{bovzivc2021mixed} dataset, with all the evaluation metrics. 

\begin{table*}[tb]
	\centering
	\caption{
		The comparison on the Average Precision (AP), Per-Region Overlap (PRO) and Pixel AUROC metrics for supervised
anomaly localization on the MVTec AD dataset. 
	}
	\label{table:supervised_mvtec_result}
		\resizebox{\textwidth}{!}{
	\begin{tabular}{lccccccc}
		\toprule
		Category     & \makecell{DevNet \cite{pang2021explainable} \NL (arXiv2021)} & \makecell{DRA \cite{ding2022catching} \NL (CVPR2022)} & \makecell{PRN \cite{zhang2022prototypical} \NL (CVPR2023)}                                    & \makecell{SemiREST \cite{li2023efficient} \NL (arXiv2023)}                                      & \makecell{CPR\cite{li_target_2023} \NL(arXiv2023)}    &\makecell{SemiREST(weak) \cite{li2023efficient} \NL(arXiv2023)}& Ours(weak)                                                                                       \\\hline
		Carpet       
		& $45.7$/$85.8$/$97.2$                                         
		& $52.3$/$92.2$/$98.2$                                  
		& $82.0$/$97.0$/$99.0$                                                                          
		& {\color{red}{$\mathbf{89.1}$}}/{\color{blue}{$\mathbf{99.1}$}}/{\color{red}{$\mathbf{99.7}$}}    
		& $88.1$/$98.9$/{\color{blue}{$\mathbf{99.6}$}}
		& {\color{blue}{$\mathbf{88.9}$}}/{\color{blue}{$\mathbf{99.1}$}}/{\color{red}{$\mathbf{99.7}$}}
		& 88.6/{\color{red}{$\mathbf{99.3}$}}/{\color{red}{$\mathbf{99.7}$}} \\
		Grid         
		& $25.5$/$79.8$/$87.9$                                         
		& $26.8$/$71.5$/$86.0$                                  
		& $45.7$/$95.9$/$98.4$                                                                          
		& $66.4$/$97.0$/{\color{blue}{$\mathbf{99.4}$}}
		& $67.3$/{\color{red}{$\mathbf{98.7}$}}/{\color{red}{$\mathbf{99.7}$}}
		& {\color{blue}{$\mathbf{71.5}$}}/98.5/{\color{red}{$\mathbf{99.7}$}}
		& {\color{red}{$\mathbf{74.1}$}}/{\color{blue}{$\mathbf{98.6}$}}/{\color{red}{$\mathbf{99.7}$}}    \\
		Leather      
		& $8.1$/$88.5$/$94.2$                                          
		& $5.6$/$84.0$/$93.8$                                   
		& $69.7$/$99.2$/$99.7$                                                                          
		& {\color{blue}{$\mathbf{81.7}$}}/{\color{red}{$\mathbf{99.7}$}}/{\color{red}{$\mathbf{99.9}$}}    
		& $78.0$/$99.5$/{\color{blue}{$\mathbf{99.8}$}}
		& {\color{red}{$\mathbf{82.0}$}}/{\color{blue}{$\mathbf{99.6}$}}/{\color{red}{$\mathbf{99.9}$}} 
		& 80.2/{\color{blue}{$\mathbf{99.6}$}}/{\color{blue}{$\mathbf{99.8}$}} \\
		Tile         
		& $52.3$/$78.9$/$92.7$                                         
		& $57.6$/$81.5$/$92.3$                                  
		& $96.5$/$98.2$/$99.6$                                                 
		& $96.9$/$98.9$/{\color{blue}{$\mathbf{99.7}$}} 
		& {\color{blue}{$\mathbf{97.2}$}}/{\color{blue}{$\mathbf{99.0}$}}/{\color{blue}{$\mathbf{99.7}$}}
		& $96.6$/$98.7$/{\color{blue}{$\mathbf{99.7}$}}
		& {\color{red}{$\mathbf{98.0}$}}/{\color{red}{$\mathbf{99.4}$}}/{\color{red}{$\mathbf{99.8}$}}    \\
		Wood         
		& $25.1$/$75.4$/$86.4$                                         
		& $22.7$/$69.7$/$82.9$                                  
		& $82.6$/$95.9$/$97.8$                                                                          
		& $88.7$/97.9/99.2
		& {\color{blue}{$\mathbf{90.7}$}}/{\color{red}{$\mathbf{98.4}$}}/{\color{red}{$\mathbf{99.5}$}} 
		& $86.2$/$97.1$/$98.6$
		& {\color{red}{$\mathbf{92.7}$}}/{\color{blue}{$\mathbf{98.3}$}}/{\color{blue}{$\mathbf{99.3}$}}   \\\hline
		Average      
		& $31.3$/$81.7$/$91.7$                                         
		& $33.0$/$79.8$/$90.6$                                  
		& $75.3$/$97.2$/$98.9$                                                                          
		& $84.7$/$98.5$/99.5
		& $84.3$/{\color{blue}{$\mathbf{98.9}$}}/{\color{blue}{$\mathbf{99.6}$}}
		& {\color{blue}{$\mathbf{85.0}$}}/$98.6$/99.5
		& {\color{red}{$\mathbf{86.7}$}}/{\color{red}{$\mathbf{99.0}$}}/{\color{red}{$\mathbf{99.7}$}}   \\\hline
		Bottle       
		& $51.5$/$83.5$/$93.9$                                         
		& $41.2$/$77.6$/$91.3$                                  
		& $92.3$/$97.0$/$99.4$                        
		& {\color{blue}{$\mathbf{93.6}$}}/{\color{red}{$\mathbf{98.5}$}}/$99.5$   
		& {\color{blue}{$\mathbf{93.6}$}}/{\color{red}{$\mathbf{98.5}$}}/{\color{blue}{$\mathbf{99.6}$}}
		& {\color{blue}{$\mathbf{93.6}$}}/{\color{blue}{$\mathbf{98.4}$}}/$99.5$
		& {\color{red}{$\mathbf{95.5}$}}/$97.3$/{\color{red}{$\mathbf{99.7}$}}    \\
		Cable        
		& $36.0$/$80.9$/$88.8$                                         
		& $34.7$/$77.7$/$86.6$                                  
		& $78.9$/{\color{red}{$\mathbf{97.2}$}}/$98.8$                                                  
		& {\color{blue}{$\mathbf{89.5}$}}/$95.9$/$99.2$  
		& $88.1$/$94.5$/{\color{blue}{$\mathbf{99.4}$}}
		& $86.5$/$96.3$/$99.3$
		& {\color{red}{$\mathbf{90.9}$}}/{\color{blue}{$\mathbf{96.8}$}}/{\color{red}{$\mathbf{99.5}$}}                           \\
		Capsule      
		& $15.5$/$83.6$/$91.8$                                         
		& $11.7$/$79.1$/$89.3$                                  
		& $62.2$/$92.5$/$98.5$                                                 
		& $60.0$/97.0/$98.8$                           
		& {\color{blue}{$\mathbf{65.8}$}}/$96.7$/{\color{red}{$\mathbf{99.4}$}}
		& $58.4$/{\color{red}{$\mathbf{97.6}$}}/{\color{blue}{$\mathbf{99.1}$}} 
		& {\color{red}{$\mathbf{65.9}$}}/{\color{blue}{$\mathbf{97.1}$}}/{\color{blue}{$\mathbf{99.1}$}}   \\
		Hazelnut     
		& $22.1$/$83.6$/$91.1$                                         
		& $22.5$/$86.9$/$89.6$                                  
		& 93.8/$97.4$/$99.7$                        
		& $92.2$/$98.3$/{\color{blue}{$\mathbf{99.8}$}}      
		& {\color{red}{$\mathbf{94.4}$}}/{\color{blue}{$\mathbf{98.7}$}}/{\color{blue}{$\mathbf{99.8}$}}
		& $86.0$/$97.3$/$99.7$
		& {\color{blue}{$\mathbf{94.2}$}}/{\color{red}{$\mathbf{99.3}$}}/{\color{red}{$\mathbf{99.9}$}}    \\
		Metal Nut    
		& $35.6$/$76.9$/$77.8$                                         
		& $29.9$/$76.7$/$79.5$                                  
		& $98.0$/$95.8$/$99.7$                                                                          
		& {\color{blue}{$\mathbf{99.1}$}}/$98.2$/{\color{red}{$\mathbf{99.9}$}}
		& 98.6/{\color{blue}{$\mathbf{98.4}$}}/{\color{blue}{$\mathbf{99.8}$}}
		& $98.3$/$98.1$/{\color{blue}{$\mathbf{99.8}$}}
		& {\color{red}{$\mathbf{99.2}$}}/{\color{red}{$\mathbf{98.9}$}}/{\color{red}{$\mathbf{99.9}$}}  \\
		Pill         
		& $14.6$/$69.2$/$82.6$                                         
		& $21.6$/$77.0$/$84.5$                                  
		& {\color{red}{$\mathbf{91.3}$}}/$97.2$/{\color{red}{$\mathbf{99.5}$}}
		& $86.1$/{\color{red}{$\mathbf{98.9}$}}/{\color{blue}{$\mathbf{99.3}$}}      
		& {\color{blue}{$\mathbf{90.7}$}}/{\color{red}{$\mathbf{98.9}$}}/{\color{red}{$\mathbf{99.5}$}}
		& $89.6$/{\color{red}{$\mathbf{98.9}$}}/{\color{red}{$\mathbf{99.5}$}}
		& $83.6$/{\color{blue}{$\mathbf{98.3}$}}/$99.0$   \\
		Screw        
		& $1.4$/$31.1$/$60.3$                                          
		& $5.0$/$30.1$/$54.0$                                   
		& $44.9$/$92.4$/$97.5$                                                 
		& {\color{blue}{$\mathbf{72.1}$}}/{\color{blue}{$\mathbf{98.8}$}}/{\color{red}{$\mathbf{99.8}$}}  
		& {\color{red}{$\mathbf{72.5}$}}/{\color{red}{$\mathbf{98.9}$}}/{\color{red}{$\mathbf{99.8}$}}
		& $67.9$/$98.6$/{\color{red}{$\mathbf{99.8}$}}
		& $71.2$/$96.8$/{\color{blue}{$\mathbf{99.6}$}}    \\
		Toothbrush   
		& $6.7$/$33.5$/$84.6$                                          
		& $4.5$/$56.1$/$75.5$                                   
		& $78.1$/$95.6$/{\color{blue}{$\mathbf{99.6}$}}                        
		& $74.2$/{\color{blue}{$\mathbf{97.1}$}}/{\color{blue}{$\mathbf{99.6}$}}                          
		& {\color{red}{$\mathbf{84.8}$}}/{\color{red}{$\mathbf{98.0}$}}/{\color{red}{$\mathbf{99.7}$}}
		& $73.3$/$96.7$/{\color{blue}{$\mathbf{99.6}$}}
		& {\color{blue}{$\mathbf{82.7}$}}/$95.4$/{\color{red}{$\mathbf{99.7}$}}    \\
		Transistor   
		& $6.4$/$39.1$/$56.0$                                          
		& $11.0$/$49.0$/$79.1$                                  
		& $85.6$/$94.8$/$98.4$                        
		& $85.5$/$97.8$/{\color{blue}{$\mathbf{98.6}$}}                           
		& {\color{red}{$\mathbf{88.1}$}}/{\color{blue}{$\mathbf{98.0}$}}/$98.4$
		& $86.4$/$97.9$/{\color{blue}{$\mathbf{98.6}$}} 
		& {\color{red}{$\mathbf{88.1}$}}/{\color{red}{$\mathbf{98.4}$}}/{\color{red}{$\mathbf{99.0}$}}   \\
		Zipper       
		& $19.6$/$81.3$/$93.7$                                         
		& $42.9$/$91.0$/$96.9$                                  
		& $77.6$/$95.5$/$98.8$                                                                          
		& $91.0$/{\color{red}{$\mathbf{99.2}$}}/{\color{blue}{$\mathbf{99.7}$}}  
		& {\color{red}{$\mathbf{91.6}$}}/98.9/{\color{red}{$\mathbf{99.8}$}}
		& {\color{blue}{$\mathbf{91.3}$}}/{\color{red}{$\mathbf{99.2}$}}/{\color{red}{$\mathbf{99.8}$}}
		& $90.5$/{\color{blue}{$\mathbf{99.1}$}}/{\color{blue}{$\mathbf{99.7}$}}   \\\hline
		Average      
		& $20.9$/$66.3$/$82.1$                                         
		& $22.5$/$70.1$/$82.6$                                  
		& $80.3$/$95.5$/$99.0$                                                                          
		& $84.3$/{\color{red}{$\mathbf{98.0}$}}/{\color{blue}{$\mathbf{99.4}$}}
		& {\color{red}{$\mathbf{86.8}$}}/{\color{blue}{$\mathbf{97.9}$}}/{\color{red}{$\mathbf{99.5}$}}
		& $83.2$/{\color{blue}{$\mathbf{97.9}$}}/{\color{red}{$\mathbf{99.5}$}}
		& {\color{blue}{$\mathbf{86.2}$}}/$97.7$/{\color{red}{$\mathbf{99.5}$}}   \\\hline
		Total Average 
		& $24.4$/$71.4$/$85.3$                                         
		& $26.0$/$73.3$/$85.3$                                  
		& $78.6$/$96.1$/$99.0$                                                                          
		& $84.4$/{\color{blue}{$\mathbf{98.1}$}}/{\color{blue}{$\mathbf{99.5}$}}
		& {\color{blue}{$\mathbf{86.0}$}}/{\color{red}{$\mathbf{98.3}$}}/{\color{red}{$\mathbf{99.6}$}}
		& $83.8$/98.1/{\color{blue}{$\mathbf{99.5}$}}
		& {\color{red}{$\mathbf{86.4}$}}/{\color{blue}{$\mathbf{98.2}$}}/{\color{red}{$\mathbf{99.6}$}}    \\
		\bottomrule
	\end{tabular}
	}
\end{table*}

\begin{table}[tb]
	
	\centering
	\caption{Supervised anomaly detection (Image AUROC) on MVTec AD \cite{bergmann2019mvtec}. Results are averaged over
		all categories. 	}
	\label{table:image_auc_result}
		\begin{tabular}{cccccc}
			\toprule
			\makecell{DevNet\cite{pang2021explainable}}&
			\makecell{DRA\cite{ding2022catching}}&
			\makecell{BGAD\cite{yao2022explicit}}&
			\makecell{PRN\cite{zhang2022prototypical}}&
			\makecell{CPR\cite{li_target_2023}}
			&\makecell{Ours(weak)}
			                       \\ \hline                                    
			$94.5$                                                    
			&$95.9$                                              
			&$99.3$
			&99.4
			&{\color{red}{$\mathbf{99.7}$}}                                               
			&{\color{blue}{$\mathbf{99.6}$}}                                                   \\
			\bottomrule
		\end{tabular}
\end{table}
\begin{table}[tb]
	\centering
	\caption{
		Supervised anomaly localization and detection performance (AP, PRO, Pixel AUROC and Image AUROC) on KolektorSDD2 \cite{bovzivc2021mixed}.}
	\label{table:ksdd2_seg_result}
\begin{tabular}{@{}lcccc@{}}\toprule
Method   & AP   & PRO  & P\_AUROC & I\_AUROC \\\midrule
PRN \cite{zhang2022prototypical}      & 72.5 & 94.9 & 97.6     & 96.4     \\
SemiREST \cite{li2023efficient} & {\color{blue}{\textbf{73.6}}} & 96.7 & 98.0     & 96.8     \\
SemiREST (weak)\cite{li2023efficient} & 72.1 & {\color{blue}{\textbf{97.5}}} & {\color{blue}{\textbf{99.1}}}     & {\color{blue}{\textbf{97.4}}}     \\
Ours (weak)     & {\color{red}{\textbf{78.4}}} & {\color{red}{\textbf{98.6}}} & {\color{red}{\textbf{99.6}}}     & {\color{red}{\textbf{98.0}}}    \\\bottomrule
\end{tabular}
\end{table}

\begin{figure}[ht]  
	\vspace{-0.1em}
  \centering{
\includegraphics [width=0.95\textwidth]{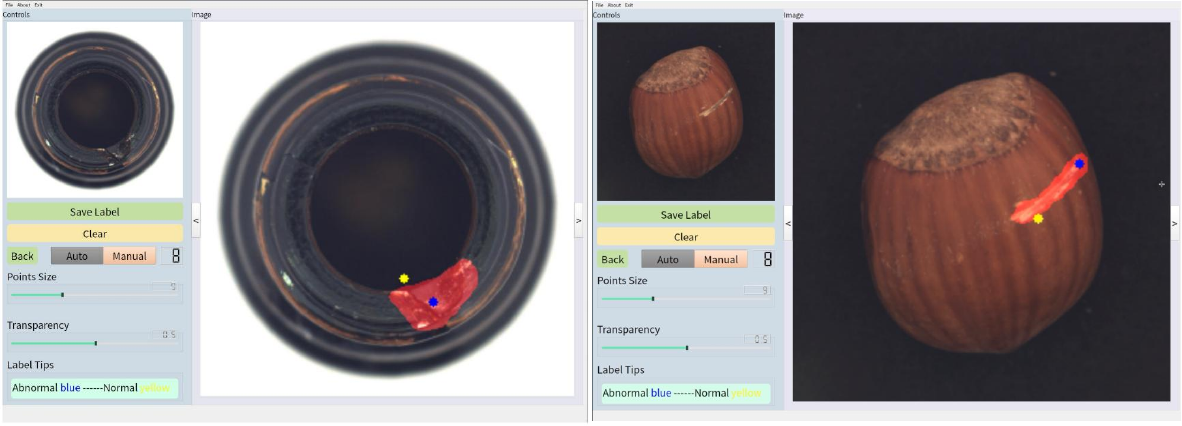}
\caption{
  Our labeling tool. The positive click (shown in blue) and negative click (shown in yellow) guide the label generation successfully. Better view in color. 
}
  \label{fig:tool}
}

\end{figure}

\subsection{Ablation study}
\label{subsec:ablation}
To obtain the insight of the module contributions,
an ablation study is conducted and the results are reported in
Tab.~\ref{table:ablation_click_result} and Tab.~\ref{table:ablation_seg_result} for
ADClick and ADClick-Seg, respectively. From Tab.~\ref{table:ablation_click_result} we can see that the language features and the fine-tuning process on the stimulated defects nearly equally contribute to the final performance of ADClick, both in terms of AP and mIoU. 
\begin{table}[tb]
\centering
	\caption{
		Ablation study results (AP/PRO/P\_AUROC/mIoU) of ADClick on MVTec-AD \cite{bergmann2019mvtec}.
	}
	\label{table:ablation_click_result}
\begin{tabular}{@{}cccc@{}}\toprule
Language & Finetune &  5-click &  3-click \\\midrule
         &          & 93.4/98.3/{\color{red}{\textbf{99.7}}}/75.5                                & 90.9/97.8/99.5/70.9                                \\
\checkmark        &          & 93.5/98.7/{\color{red}{\textbf{99.7}}}/75.6                                & 91.2/{\color{red}{\textbf{98.4}}}/{\color{red}{\textbf{99.6}}}/71.2                                \\
         & \checkmark        & 94.0/98.5/{\color{red}{\textbf{99.7}}}/77.0                                & 91.8/97.9/{\color{red}{\textbf{99.6}}}/72.8                                \\
\checkmark        & \checkmark        & {\color{red}{\textbf{94.1}}}/{\color{red}{\textbf{98.8}}}/{\color{red}{\textbf{99.7}}}/{\color{red}{\textbf{77.7}}}                                & {\color{red}{\textbf{92.1}}}/{\color{red}{\textbf{98.4}}}/{\color{red}{\textbf{99.6}}}/{\color{red}{\textbf{73.7}}} \\\bottomrule
\end{tabular}
\end{table}

On the other hand, the model pertaining on the other subcategories (Pretrain), the IIS-like training style (IIS), the language prompts (Lang.) and the Label Abandon strategy of training (Abandon) are considered contributing modules for ADClick-Seg. We can see that the pretraining stage and language guidance contribute similarly to the segmentation model. The IIS-like training style lifts the AD performance only when language guidance is involved. The label abandon strategy lead to a slightly better AD performance compared with the model trained without it.  

\begin{table}[tb]
 \centering
	\caption{
		Ablation study results of ADClick-Seg on MVTec-AD \cite{bergmann2019mvtec}.
	}
	\label{table:ablation_seg_result}
\begin{tabular}{@{}cccccccc@{}}
\toprule
Pretrain & IIS & Lang. & Abandon & AP & PRO &P\_AUROC &I\_AUROC \\\midrule
\checkmark &          & & &84.4&{\color{red}{\textbf{98.2}}}&\underline{99.4}&{\color{blue}{\textbf{99.6}}} \\
& \checkmark           &          &          &80.4&\underline{97.9}&99.2&99.2\\
& & \checkmark & &84.2&{\color{red}{\textbf{98.2}}}&\underline{99.4}&{\color{blue}{\textbf{99.6}}}\\
\checkmark       & \checkmark           &          && 81.8&95.4 &99.0 &99.4  \\
\checkmark       &              & \checkmark       && 85.0 &{\color{red}{\textbf{98.2}}} &\underline{99.4}&{\color{red}{\textbf{99.7}}}    \\
& \checkmark           & \checkmark       & &\underline{86.0}&{\color{red}{\textbf{98.2}}} &{\color{blue}{\textbf{99.5}}} &{\color{red}{\textbf{99.7}}}  \\
\checkmark       & \checkmark           & \checkmark       & &{\color{red}{\textbf{86.7}}} &97.8 & {\color{blue}{\textbf{99.5}}} & {\color{red}{\textbf{99.7}}} \\
\checkmark       & \checkmark           & \checkmark       &  \checkmark  &{\color{blue}{\textbf{86.4}}} &{\color{red}{\textbf{98.2}}}  & {\color{red}{\textbf{99.6}}} & {\color{blue}{\textbf{99.6}}} \\
\bottomrule
\end{tabular}
\end{table}

\subsection{The annotation tool}

Based on ADClick, we develop an interactive annotation tool that will be available to the
public soon. Users can easily label their anomaly data smoothly using our tool, as shown
in Fig.~\ref{fig:tool}. A demonstration video of this tool can also be found in the
supplementary of this paper. 

\section{Conclusion}
\label{sec:conclusion}
This work addresses the problem of efficient anomaly labeling to reduce the cost of manual
pixel annotations for anomaly detection. We propose a sophisticated Swin transformer model
that can generate high-quality pixel labels of anomalous regions based on the cross-modal
inputs including human clicks, language prompts and deep feature residuals. To the best of
our knowledge, it is the first time that the interactive segmentation method has been adopted to
generate training labels for AD tasks. In addition, the proposed algorithm is multi-functional: new SOTA performances of weakly-supervised AD are obtained by using the
ADClick model fine-tuned on the training data with pseudo labels. This work shed light on the
fundamental differences between the label generation in the big-data scenario and that for
the AD tasks with limited samples. More efficient AD labels and better AD results could be
achieved via further exploration into the distinct nature of AD supervision.

\bibliographystyle{splncs04}
\bibliography{anomaly_detection,interactive_segmentation,others}

\end{document}